\documentclass[lettersize,journal]{IEEEtran}
\usepackage{amsmath,amsfonts}
\usepackage{algorithmic}
\usepackage{algorithm}
\usepackage{array}
\usepackage[caption=false,font=normalsize,labelfont=sf,textfont=sf]{subfig}
\usepackage{textcomp}
\usepackage{stfloats}
\usepackage{url}
\usepackage{verbatim}
\usepackage{graphicx}
\usepackage{cite}

\usepackage{lineno,hyperref}
\usepackage{multirow}
\usepackage{booktabs}
\usepackage{diagbox}
\usepackage{makecell}
\usepackage[normalem]{ulem}
\useunder{\uline}{\ul}{}

\begin{document}

\title{FoME: A Foundation Model for EEG using Adaptive Temporal-Lateral Attention Scaling}

\author{Enze Shi, Kui Zhao, Qilong Yuan, Jiaqi Wang, Huawen Hu, Sigang Yu, Shu Zhang$^{*}$
\thanks{Under Review.}
\thanks{Enze Shi, Kui Zhao, Qilong Yuan, Jiaqi Wang, Huawen Hu, Sigang Yu, and Shu Zhang$^{*}$ are with the Center for Brain and Brain-Inspired Computing Research, School of Computer Science, Northwestern Polytechnical University, Xi'an 710072, China. (*Corresponding author)}}

\markboth{Journal of \LaTeX\ Class Files,~Vol.~14, No.~8, August~2021}%
{Shell \MakeLowercase{\textit{et al.}}: A Sample Article Using IEEEtran.cls for IEEE Journals}


\maketitle

\begin{abstract}
Electroencephalography (EEG) is a vital tool to measure and record brain activity in neuroscience and clinical applications, yet its potential is constrained by signal heterogeneity, low signal-to-noise ratios, and limited labeled datasets. In this paper, we propose FoME (Foundation Model for EEG), a novel approach using adaptive temporal-lateral attention scaling to address above-mentioned challenges. FoME is pre-trained on a diverse 1.7TB dataset of scalp and intracranial EEG recordings, comprising 745M parameters trained for 1,096k steps. Our model introduces two key innovations: a time-frequency fusion embedding technique and an adaptive time-lateral attention scaling (ATLAS) mechanism. These components synergistically capture complex temporal and spectral EEG dynamics, enabling FoME to adapt to varying patterns across diverse data streams and facilitate robust multi-channel modeling. Evaluations across four downstream tasks demonstrate FoME's superior performance in classification and forecasting applications, consistently achieving state-of-the-art results. To conclude, FoME establishes a new paradigm for EEG analysis, offering a versatile foundation that advances brain-computer interfaces, clinical diagnostics, and cognitive research across neuroscience and related fields. Our code will be available at https://github.com/1061413241/FoME.
\end{abstract}


\section{Introduction}
\IEEEPARstart{E}{EG} (electroencephalography), as a cost-effective neurophysiological monitoring technique, holds significant research and application potential across cognitive neuroscience, psychiatry, and clinical diagnostics \cite{lotte2018review}. Its high temporal resolution enables near real-time reflection of dynamic brain patterns, offering valuable insights into cerebral mechanisms \cite{lahane_review_2019}. Both invasive and non-invasive EEG acquisition methods have found widespread application in brain-computer interfaces \cite{zhang2023differentiating}, sleep monitoring \cite{phan2021xsleepnet}, epilepsy detection \cite{kovacs2021cost}, and emotion recognition \cite{yi2024learning}, among other domains.

Extant research has primarily focused on addressing specific tasks, such as the MEET model for emotion decoding \cite{shi2023meet}, EEG Conformer for motor imagery classification \cite{song2022eeg}, and SEEG-Net for epileptic seizure classification \cite{wang2022seeg}. While these studies have yielded substantial progress, they typically require de novo model training and often rely on large volumes of labeled data, rendering large-scale signal annotation financially impractical \cite{diachenko2022improved}. Moreover, the inherent heterogeneity and low signal-to-noise ratio of EEG pose challenges to model generalization and transfer capabilities, impeding further exploration of EEG's potential.

The success of foundation models in natural language processing and computer vision domains has presented a compelling paradigm: self-supervised training on vast datasets to learn abstract representations, followed by fine-tuning with limited labeled data \cite{vaswani2017attention, dosovitskiy2020image, brown2020language}. This approach offers a powerful solution to the aforementioned limitations in EEG research, underscoring the urgent need for an EEG foundation model. However, developing an EEG foundation model from scratch presents several challenges:

First of all, different studies often employ diverse acquisition systems, resulting in significant differences in sampling rates, electrode positions, and quantities. Despite the standardization efforts of the International 10-20 system, strict standardization of EEG acquisition protocols is still difficult to achieve due to variations in experimental personnel, environments, and equipment. For instance, the DEAP dataset uses a 512 Hz sampling rate with 32 active AgCl electrodes arranged according to the 10-20 system \cite{koelstra2011deap}, while a visual object recognition dataset employs a 64-channel EASYCAP system with a 1000 Hz sampling rate \cite{gifford2022large}. These differences pose challenges in the data processing phase of EEG foundation model construction.

Secondly, EEG signals exhibit distinct physiological characteristics across various scenarios. In motor imagery tasks, for example, the temporal features of EEG signals may be highly task-correlated due to short duration \cite{zhang2022explainable}. In contrast, epilepsy detection requires modeling of long-term temporal signals, potentially confounded by limb movement artifacts \cite{chen2022brainnet}. In emotion analysis or sleep monitoring tasks, specific frequency band information may be more conducive to model discrimination \cite{zheng2018emotionmeter}. Optimally leveraging both temporal and spectral information is thus crucial.

Finally, the brain's role as the neural hub controlling and reflecting various behavioral and physiological indices necessitates a model with multifaceted capabilities. These include: (1) comprehensive EEG modeling to capture signal periodicity and trends, enabling both short-term and long-term sequence forecasting; (2) serve as a component in various classification or detection tasks (e.g., epilepsy detection, sleep quality monitoring, emotion classification); and (3) out-of-the-box functionality, unrestricted by specific datasets or tasks, achieving high performance through simple fine-tuning and ideally demonstrating zero-shot capabilities.

\begin{figure}[t]
\centering
\includegraphics[width=1.0\columnwidth]{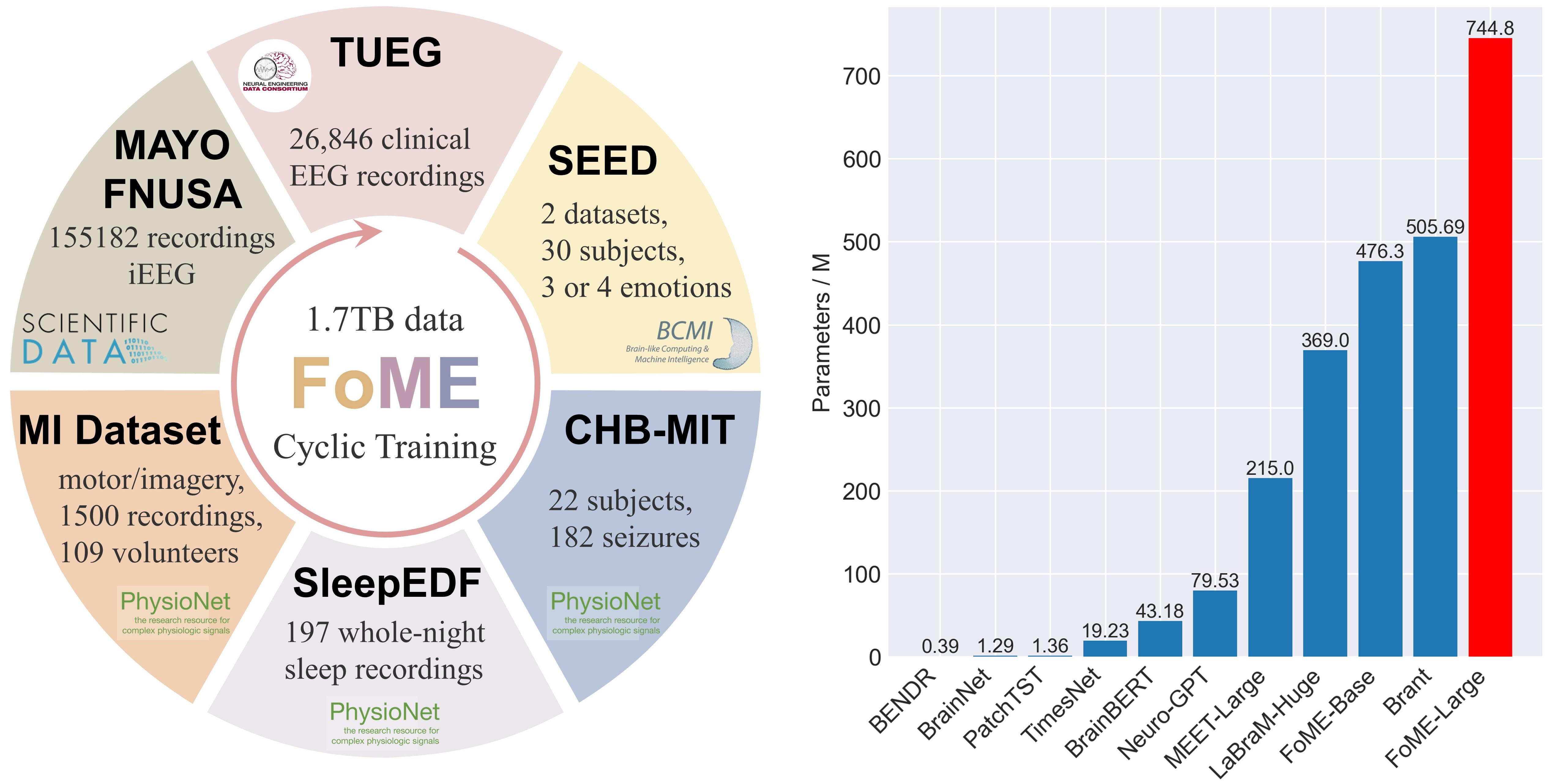}
\caption{Left: FoME's large-scale heterogeneous pre-training dataset; Right: Parameter scale of existing EEG correlation models.}
\label{intro}
\end{figure}

To address these challenges, we propose FoME: a foundation model for EEG using adaptive temporal-lateral attention scaling. This model aims to bridge the existing gap by developing a large-scale, versatile foundation model capable of effectively handling diverse EEG datasets, encompassing both scalp and intracranial recordings. By leveraging self-supervised large-scale pre-training and efficient architectural design, FoME seeks to establish a new benchmark for universal modeling of brain signals, paving the way for widespread applications in brain-computer interfaces, clinical neuroscience, and beyond. Our contributions are summarized in three folds:

\begin{itemize}
    \item {We propose a novel EEG Foundation Model (FoME) pre-trained on a massive and diverse dataset comprising over 1.7TB of heterogeneous and multimodal data, as shown in Fig. \ref{intro}. Trained for 1,096k steps with 745M parameters, FoME establishes a robust foundation for various EEG-based applications.}
    \item {To effectively capture the intricate dynamics of EEG signals, we propose time-frequency fusion embedding and adaptive time-lateral attention scaling (ATLAS) mechanism. Our approach integrates temporal and spectral information into a unified representation, enabling the model to learn discriminative features at multiple scales. ATLAS dynamically adapts to the changing temporal and spatial patterns across data, facilitating accurate and robust multi-channel modeling.}
    \item {Through extensive experiments on four challenging downstream tasks based on multiple datasets, we demonstrate FoME's superior performance in classification and forecasting, achieving state-of-the-art results. Our findings demonstrate a more advanced advantage of FoME in signal forecasting over other methods, supported by rigorous statistical and visual analysis, underscore the effectiveness of our pre-training strategy and highlight the transformative potential of FoME in critical applications such as disease surveillance, early warning systems, and personalized healthcare.}
\end{itemize}

\section{Related Work}
The development of large-scale, generalizable models for brain signals, particularly EEG data, has been a burgeoning area of research in recent years. Our work on the FoME builds upon advancements in self-supervised learning, transfer learning across datasets, and the emerging field of ``brain foundation models".

\subsection{Self-supervised Pre-training for EEG Signals}
The remarkable success of self-supervised pre-training techniques in natural language processing and computer vision has motivated their exploration in the domain of brain signal analysis \cite{devlin2018bert}. These methods leverage the abundance of unlabeled data to learn robust and transferable representations, which can then be fine-tuned for specific downstream tasks.

In the EEG domain, several studies have investigated self-supervised learning approaches. BENDR \cite{kostas2021bendr} adapted the Wav2Vec 2.0 framework \cite{baevski2020wav2vec} to learn compressed representations of raw EEG signals using contrastive learning. Banville et al. explored temporal context prediction and contrastive predictive coding as self-supervised pretext tasks for EEG representation learning, demonstrating their effectiveness for sleep staging and pathology detection \cite{banville2021uncovering}. ContraWR \cite{yang2023self} used global statistics to distinguish EEG signals associated with different sleep stages through a contrastive learning approach. While these studies have shown the promise of self-supervised learning for EEG data, they have primarily focused on specific BCI tasks or have been limited to small-scale datasets and models. The potential of self-supervised pre-training on large-scale, diverse EEG datasets remains largely unexplored, which is a key focus of our work on the FoME model.

\subsection{Transfer Learning for Diverse EEG Datasets}
A key challenge in EEG research is the diversity of datasets, which can vary significantly in terms of electrode configurations, sampling rates, and experimental protocols. Addressing this heterogeneity is crucial for developing robust and generalizable models.

Recent studies have explored transfer learning techniques to bridge the gap between disparate EEG datasets. Han et al. combined graph neural networks and transfer learning to perform motor imagery EEG decoding across datasets with different electrode montages \cite{han2023eeg}. Gu et al. proposed a two-network architecture that could learn from both shared and complete channel information, enabling coherent performance improvements when transferring between datasets \cite{gu2023generalizable}. MMM introduced a pre-training framework that learns topology-agnostic representations, enabling the model to handle EEG data from different electrode configurations \cite{yi2024learning}.

\subsection{Emergence of ``Brain Foundation Models"}
The concept of ``foundation models", which has gained significant traction in natural language processing and computer vision, is now being explored in the domain of brain signal analysis. These large-scale, pre-trained models can be efficiently fine-tuned for a variety of downstream tasks, promising increased efficiency and generalization compared to task-specific models.

In the context of intracranial recordings, Wang et al. proposed BrainBERT, an embeddings-based model for stereo-electroencephalographic (SEEG) data \cite{wang2023brainbert}. BrainBERT utilizes self-supervised pre-training on SEEG spectrograms to generate versatile representations. Zhang et al. introduced Brant, a large-scale foundation model specifically designed for SEEG data, which can be adapted for various intracranial signal processing tasks \cite{zhang2024brant}. Jiang et al. proposed LaBraM, which uses neural tokenizer to encode EEG fragments into discrete neural codes and pre-train neural Transformers in the form of predictive masks \cite{jiang2024large}.

% \textbf{Lack of multimodal learning of brain signals}

Previous studies have often aimed to construct unified topological maps to represent diverse channel patterns, or employed fixed channel encodings to differentiate between electrodes. Departing from these approaches, we delve into the multi-scale characteristics of EEG signals to elucidate intricate electrode features. Consequently, we propose a novel adaptive temporal-lateral attention scaling (ATLAS) mechanism. ATLAS is designed to flexibly scale attention in both temporal and spatial dimensions, thereby more effectively capturing features and patterns within structurally varied data. As a result, our model is no longer confined to specific EEG datasets, obviating the need for uniform topological rules or custom channel encodings for each dataset. The proposed FoME can adaptively model multiple electrode structures, serving as a foundation for task-agnostic and data-agnostic EEG signal analysis.

\begin{figure*}[t]
\centering
\includegraphics[width=1.0\textwidth]{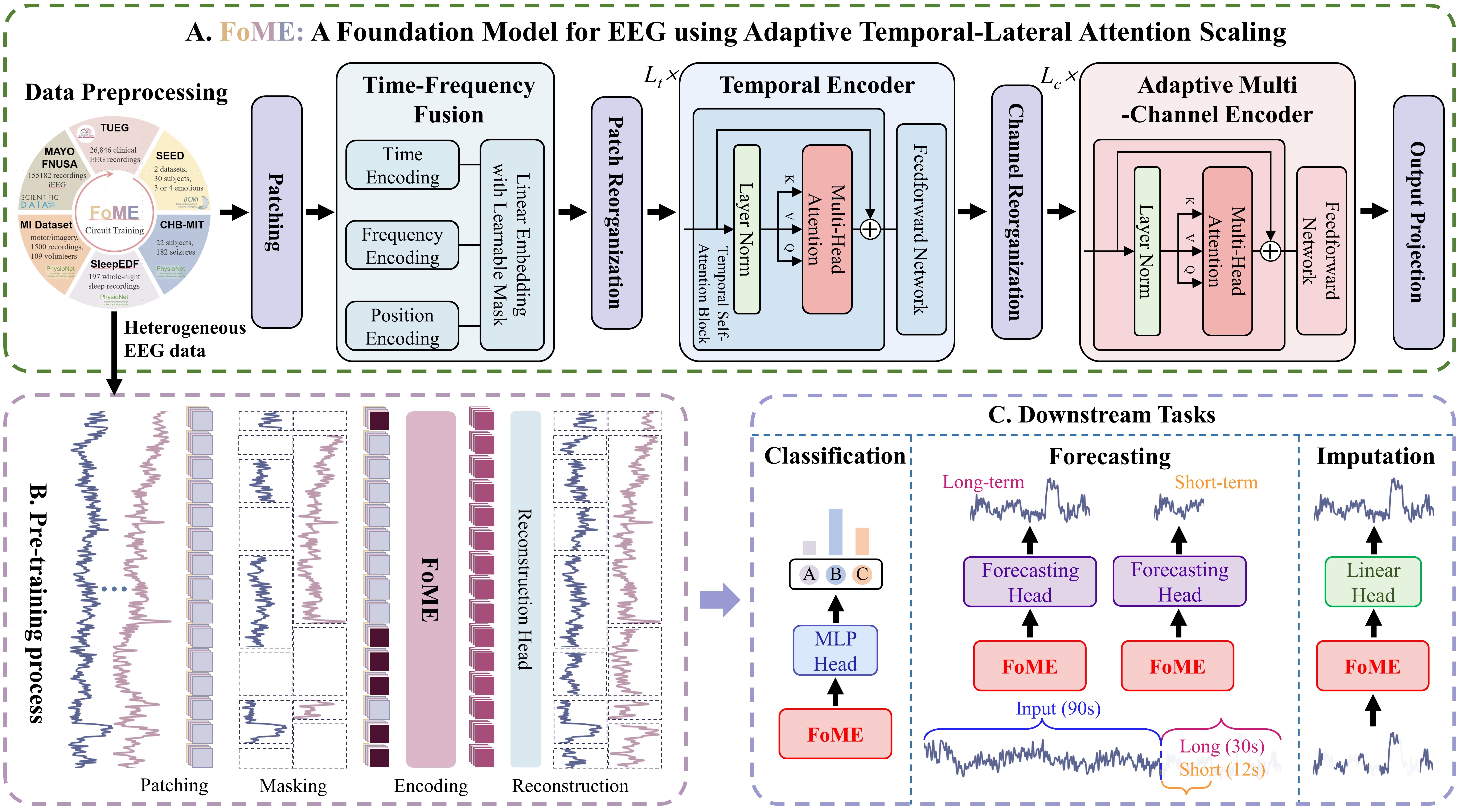}
\caption{Overview of the FoME architecture and workflow. (A) Raw EEG signal processing and key components of FoME. (B) Self-supervised learning approach using masked signal reconstruction. (C) Fine-tuning process for various downstream EEG analysis tasks.}
\label{overview}
\end{figure*}

% 方法
\section{Method}
\label{sec:method}

In this section, we provide a comprehensive overview of the proposed FoME architecture and pipeline in Section \ref{sec:overview}, followed by a detailed description of the model's key designs in Section \ref{sec:architecture}. Subsequently, Sections \ref{sec:pretrain} and \ref{sec:finetune} elaborate on FoME's pre-training strategy, leveraging a vast corpus of heterogeneous datasets, and its fine-tuning approaches for diverse downstream tasks.

\subsection{Overview}
\label{sec:overview}

As illustrated in Fig. \ref{overview}-A, the model processes raw EEG signals from diverse datasets, which are preprocessed and segmented into non-overlapping patches. These patches are sequentially fed into FoME, which comprises three primary components: a time-frequency fusion module, a temporal encoder, and an adaptive multi-channel encoder. The time-frequency fusion module integrates temporal, frequency, and positional information of each patch into a unified embedding. The temporal encoder captures temporal dependencies along the time dimension, while the adaptive multi-channel encoder learns latent associations between different channels, modeling a global channel attention network. These components collaboratively process raw EEG data and learn meaningful representations that capture both the temporal dynamics and spatial relationships inherent in brain activity.

The model adopts a self-supervised learning paradigm, employing a masked signal reconstruction task to train on extensive unlabeled EEG data. As depicted in Fig. \ref{overview}-B, the multi-channel signal sequence is partitioned into fixed-length subsequences, termed patches.  We replace the embeddings of randomly selected channel and positional patches with learnable mask embeddings [MASK]. The FoME backbone then reconstructs the original input signal sequence through a lightweight linear reconstruction head. Fig. \ref{overview}-C illustrates the fine-tuning methodology and broad applications of the model for various downstream tasks. By replacing the corresponding output projection head according to the specific task and signal input, FoME can be adapted for EEG signal classification, disease detection, short-term and long-term signal prediction, and imputation.

\subsection{Foundation Model Architecture}
\label{sec:architecture}

\subsubsection{Patching}

Prior to feeding data into the model, we segment the continuous EEG signals into non-overlapping patches. This approach effectively addresses the challenges associated with processing long, continuous EEG recordings while preserving the ability to capture both local and global temporal patterns. Consequently, the model can process EEG data at multiple scales. Specifically, the preprocessed EEG and iEEG data are represented as $X_i\in\mathbb{R}^{(N\times C\times T)}, i\in\{1,2,...,M\}$, where $M$ denotes the total number of datasets, $N$ is the number of samples, $C$ is the number of channels (electrodes), and $T$ is the total number of time points. We divide the continuous EEG signals into non-overlapping segments of length $L$, resulting in patches: $P_i\in\mathbb{R}^{(C\times (N\times F)\times L)}$, where $F=\lfloor T/L\rfloor$ represents the number of patches obtained from each sample.

\begin{figure*}[t]
\centering
\includegraphics[width=1.0\textwidth]{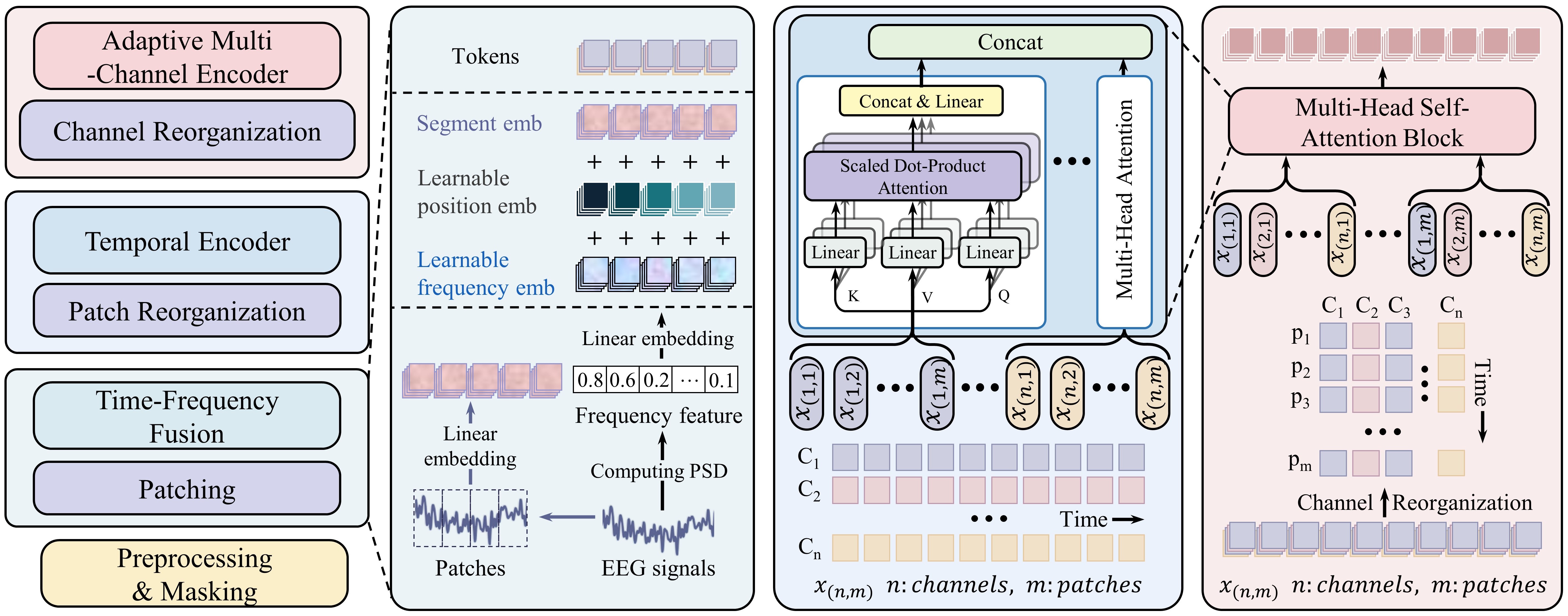}
\caption{Detailed structural presentation of time-frequency fusion encoding, temporal encoder and adaptive multi-channel encoder.}
\label{method}
\end{figure*}

\subsubsection{Time-Frequency Fusion Encoding}

It is designed to comprehensively capture both temporal and spectral characteristics inherent in EEG signals. To achieve this, we integrate time-domain and frequency-domain features into a unified representation. As illustrated in Fig. \ref{method}, in addition to segmenting the raw signal, we also extract the power density value corresponding to each frequency and calculated the power sum within the frequency band according to the preset n frequency band intervals $[f_i, f_{i+1}]$ ($i\in\{1,2,...,M\}$, where 0Hz $< f_i <$ Nyquist frequency) \cite{landau1967sampling}. Inspired by \cite{zhang2015low} and \cite{zhang2024brant}, we divide the frequency spectrum into eight bands: $\delta$ (1-4Hz), $\theta$ (4-8Hz), $\alpha$ (8-13Hz), $\beta$ (13-30Hz), $\gamma1$ (30-50Hz), $\gamma2$ (50-70Hz), $\gamma3$ (70-90Hz), and $\gamma4$ (90-100Hz). The detailed calculation process is as follows:

For a given patch $P(t)$, we first apply the Fourier transform \cite{bracewell1989fourier} to convert the time-domain signal into the frequency domain. The power spectral density, denoted as $P(f)$, is then calculated by squaring the magnitude of the frequency components:

\begin{equation}
P(f)=\frac{1}{T}\left|\mathcal{F}\{P(t)\}\right|^2=\frac{1}{T}\left| \int_{-\infty}^\infty P(t)e^{-i2\pi ft}dt \right|^2
\end{equation}
where $f$ is the frequency, $T$ is the total duration of the signal segment, and $F$ is the Fourier transform. Subsequently, we compute the sum of power within each sub-band $[f_i, f_{i+1}]$ and apply a logarithm to obtain $Power_\text{band}$. This normalization compresses the numerical range and promotes a more normal distribution of power values, facilitating model learning:

\begin{equation}
{Power}_\text{band}=\log_{10}(\sum_{f=f_{i}}^{f_{i+1}}P(f)+1)
\end{equation}

The $Power_\text{band}$ is then passed through a Softmax function and a linear embedding to obtain the frequency embedding $E_\text{freq}$: 

\begin{equation}
E_\text{freq} = \text{Linear}(\text{Softmax}(Power_\text{band})).
\end{equation}

Concurrently, we employ a linear encoder to map the patches to a latent space, resulting in the embedding $E_\text{patch}$. And a learnable positional embedding vector $E_\text{pos}$ is introduced. The final input encoding $E_\text{input}$ is obtained as: 

\begin{equation}
\begin{aligned}
E_\text{input}&=E_\text{patch}+E_\text{freq}+E_\text{pos}\\E_\text{patch}&,E_\text{freq},E_\text{pos}\in \mathbb{R}^{C\times P\times D}
\end{aligned}
\end{equation}
where $P$ is the number of patches and $D$ is the model dimension.

\subsubsection{Temporal Encoder}

The Temporal Encoder is designed to capture long-range dependencies and intricate temporal patterns within EEG data. Leveraging the success of transformer architectures in sequential modeling, this component is tailored to accommodate the unique characteristics of EEG signals. As depicted in Fig. \ref{method}, the temporal encoder consists of a stack of dense Transformer encoding layers. Each layer encompasses three key modules: multi-head self-attention, feed-forward neural network, and layer normalization with residual connections. Given an input embedding $E_\text{input}\in \mathbb{R}^{C\times P\times D}$ obtained from the previous step, we apply a trainable linear transformation along the temporal dimension to derive the query, key, and value matrices for the attention operation, denoted as $Q_\text{time}^i$, $K_\text{time}^i$, and $V_\text{time}^i$ $\in \mathbb{R}^{P\times D}$, respectively. The temporal dependencies are then modeled by computing the attention score $\text{Attn}_\text{time}^i$:

\begin{equation}
\text{Attn}_{\text{time}}^i=\text{Softmax}(Q_{\text{time}}^i(K_{\text{time}}^i)^T/\sqrt{D})V_{\text{time}}^i
\end{equation}

In the multi-head self-attention mechanism, the query, key, and value matrices are linearly projected into $D_k$, $D_k$, and $D_v$ dimensions. Subsequently, the attention function is applied to each projected version, yielding an output of dimension $D_v$. These outputs are concatenated and mapped back to the original dimension $D$ to obtain the final attention value:

\begin{equation}
\begin{aligned}\mathrm{MultiHead}(Q,K,V)&=\mathrm{Concat}(\mathrm{head}_1,...,\mathrm{head}_\mathrm{h})W^O\\\mathrm{where~head_i}&=\text{Attn}(QW_i^Q,KW_i^K,VW_i^V)\end{aligned}
\end{equation}
where $W_i^Q\in\mathbb{R}^{D\times D_k},W_i^K\in\mathbb{R}^{D\times D_k},W_i^V\in\mathbb{R}^{D\times D_v}$, and $W^O\in{\mathbb{R}^{hD_v\times D}}$ denote the projection matrix.

\subsubsection{Adaptive Multi-channel Encoder}

The adaptive multi-channel encoder is specifically designed to capture spatial relationships and inter-channel dynamics within EEG data. Recognizing the intricate interactions between different brain regions represented by distinct EEG channels, this encoder adapts to learn channel-specific characteristics and their interdependencies. The architecture of the adaptive multi-channel encoder mirrors that of the Temporal Encoder, as illustrated in Fig. \ref{method}. Given the embedding $E_\text{time}\in \mathbb{R}^{P\times D}$ from the previous step, we decouple the channel dimension from the sample dimension, resulting in $E_\text{time\_dec}\in \mathbb{R}^{C\times P\times D}$. The data is then rearranged, and a trainable linear transformation is applied along the channel dimension to obtain the query, key, and value matrices for each patch, denoted as $Q_\text{channel}^i$, $K_\text{channel}^i$, and $V_\text{channel}^i$ $\in \mathbb{R}^{C\times D}$. Finally, we compute the attention score $Attn_\text{channel}^i$, which involves the interactions between channels and represents the spatial dynamics of the brain:

\begin{equation}
\text{Attn}_{\text{channel}}^i=\text{Softmax}(Q_{\text{channel}}^i(K_{\text{channel}}^i)^T/\sqrt{D})V_{\text{channel}}^i
\end{equation}

In conclusion, heterogeneous data provide spatio-temporal characteristics of different structures contained in the signals of different tasks. By integrating time-frequency fusion, temporal encoding, and adaptive multi-channel processing, FoME effectively captures the complexity and multiscale properties of brain activity as reflected in EEG signals.

\subsection{Pre-training Using Heterogeneous EEG Datasets}
\label{sec:pretrain}

\subsubsection{Preprocessing}

Given the heterogeneity of EEG datasets, characterized by varying data structures, sampling rates, and electrode configurations, even within the same dataset, data from different subjects may be recorded using different acquisition devices. Thanks to FoME's robust support for heterogeneous data, we can simply partition data based on their structural differences, ensuring that signals within each training group exhibit consistent characteristics. The specific preprocessing pipeline is as follows:

Firstly, a notch filter is applied at 50 Hz or 60 Hz to eliminate power line interference. Subsequently, a band-pass filter is applied between 0.5 Hz and 100.5 Hz to remove high-frequency noise. The data is then resampled to 250 Hz to enhance computational efficiency. To eliminate any linear or constant offsets that might bias the analysis, a detrending process is applied. Finally, the continuous EEG data is segmented into non-overlapping windows, each containing 1500 time points (6s). Exponential moving standardization is applied to each window, normalizing the signal by removing the moving average and scaling by the moving standard deviation. For the given EEG data $X\in\mathbb{R}^{C\times T}$, the calculation formulas for the Exponential Moving Average (EMA) and the Exponential Moving Standard Deviation (ESD) are as follows:

\begin{equation}
\begin{aligned}
\mathrm{EMA}_t&=\alpha\cdot X_t+(1-\alpha)\cdot\mathrm{EMA}_{t-1}\\
\mathrm{ESD}_t&=\sqrt{\alpha\cdot(X_t-\mathrm{EMA}_t)^2+(1-\alpha)\cdot\mathrm{ESD}_{t-1}^2}\\
X'&=\frac{X-\mathrm{EMA}}{\mathrm{ESD}+\epsilon}
\end{aligned}
\end{equation}
where $\alpha$ is the smoothing factor, $X_t$ is the signal value at the time point $t$, and $X'$ is the normalized data.

\subsubsection{Pre-training Process}

Inspired by \cite{nie2022time, he2022masked, devlin2018bert}, we employ a masked signal reconstruction task for FoME pre-training, as illustrated in Fig. \ref{overview}-B. After preprocessing all datasets involved in pre-training, we further organize the data into multiple signal blocks with approximately equal data volumes. The model treats each signal block as the minimal input unit. This enables cyclical pre-training on multiple datasets. We begin by segmenting the input signal into patches and randomly replacing 40\% of the patches with learnable mask embeddings [MASK]. The entire dataset is then fed into FoME to learn latent signal representations, and a linear mapping is used to reconstruct the original input sequence. The mean squared error between the original and reconstructed signals serves as the final pre-training loss.

\subsection{Fine-tuning on Downstream Tasks}
\label{sec:finetune}

FoME can be seamlessly applied to various EEG analysis tasks, as depicted in Fig. \ref{overview}-C. In this paper, we demonstrate its utility on four practical tasks: classification, short-term forecasting, long-term forecasting, and imputation. For classification, we replace the reconstruction head with a classification head, using a Multi-Layer Perceptron (MLP) to reduce the dimensionality of the FoME output embeddings three times before applying softmax to obtain multi-class labels. For prediction tasks with a horizon of H, we replace the reconstruction head with a forecasting head, flattening the N-dimensional embeddings of size D output by FoME into an N×D dimensional vector, and then projecting it onto an H-dimensional signal sequence using a linear projection. For imputation tasks, we retain the linear reconstruction head.

\section{Experiments}

\subsection{Pre-training Setup}

\subsubsection{Model Variants}
We developed two configurations of FoME: FoME-Base and FoME-Large. The Base variant contains 476.3 million parameters, while the Large variant comprises 744.8 million parameters. Both configurations incorporate 12 temporal encoder layers and 4 adaptive multi-channel encoder layers. The primary distinction lies in the hidden layer dimensions: FoME-Base employs a feedforward network (FFN) with a dimension of 3072, whereas FoME-Large uses an FFN dimension of 7168. In accordance with established scaling laws for large models \cite{kaplan2020scaling}, we initially evaluated FoME-Base on smaller-scale datasets to extrapolate the performance of FoME-Large on larger datasets. This approach facilitated the preliminary estimation of training outcomes and informed the selection of hyperparameters and training strategies. Unless otherwise specified, all evaluation results presented in this paper are from FoME-Large.

\subsubsection{Datasets}
The pre-training corpus for FoME was derived from eight diverse datasets: TUEG \cite{harati2014tuh}, SEED \cite{zheng2015investigating}, SEED-IV \cite{zheng2018emotionmeter}, CHB-MIT \cite{chbmit2010, shoeb2009application}, Sleep-EDFx \cite{kemp2000analysis}, MI-Dataset \cite{schalk2004bci2000}, MAYO \cite{nejedly2020multicenter}, and FNUSA \cite{nejedly2020multicenter}. Among these, MAYO and FNUSA contain invasive EEG signals, while the remainder comprise scalp EEG recordings. The aggregate dataset encompasses over 30,000 recordings from more than 15,000 subjects, totaling approximately 26,000 hours of data and exceeding 1.7TB in volume. The six scalp EEG datasets exhibit sampling rates ranging from 100 Hz to 1024 Hz and incorporate over 40 distinct electrode configurations, with channel counts varying from 3 to 64. The two invasive EEG datasets feature sampling rates of 32 kHz (MAYO: 16-76 channels) and 25 kHz (FNUSA: 192 channels), respectively. The age distribution of subjects spans from 1 to 90 years. To construct the pre-training corpus, we sampled varying proportions of data from each dataset based on task requirements and data volume: 100\% of TUEG, 80\% of SEED, CHB-MIT, Sleep-EDFx, and MI-Dataset, and 50\% of MAYO and FNUSA. The remaining data were reserved for downstream signal forecasting tasks.

\subsubsection{Pre-training Methodology}
We concatenated contiguous signal segments from the pre-training data to form signal blocks ranging from 5 to 10 GB in size. Following preprocessing and signal segmentation, each signal block served as an input unit for the model, with one complete cycle through all blocks constituting an epoch. During pre-training, each input sample comprised 15 patches, each containing 1500 time points (6s). 40\% of these patches were randomly replaced with learnable mask embeddings [MASK]. We utilized a minibatch size of 12 input samples. FoME was optimized using the AdamW optimizer with $\beta_{1}=0.9$, $\beta_{2}=0.99$, and $eps=10^{-6}$. A weight decay of $1\times 10^{-2}$ was applied to stabilize the training process and mitigate overfitting. The learning rate schedule incorporated an initial rate of $2\times 10^{-6}$, which was linearly increased to a peak of $5\times 10^{-5}$ over 10,960 steps, followed by a cosine decay to $5\times 10^{-9}$ over 1,096k steps. We employed gradient accumulation to simulate larger batch sizes, updating parameters every four steps. FoME was pre-trained for 350 hours on 6 NVIDIA RTX 4090 GPUs (24GB each).

\subsection{Downstream Tasks Setup}

To assess the generalizability of FoME, we conducted extensive experiments across four critical downstream tasks: seizure classification, seizure detection, emotion recognition, sleep stage classification, short-term forecasting, long-term forecasting, and signal imputation. Table \ref{benchmark} provides a comprehensive summary of the benchmarks.

% \usepackage{multirow}
\begin{table}[h]
\caption{Summary of Experiment Benchmarks}
\centering
\begin{tabular}{clc}
\toprule
Tasks                                                        & Benchmarks                                                                        & Metrics                                                                \\ \midrule
Classification                                               & \begin{tabular}[c]{@{}l@{}}MAYO, FNUSA, SEED, \\ SleepEDFx, TUEV\end{tabular}     & \begin{tabular}[c]{@{}c@{}}Accuracy, \\ Precision, Recall\end{tabular} \\ \midrule
\begin{tabular}[c]{@{}c@{}}Seizure \\ Detection\end{tabular} & \begin{tabular}[c]{@{}l@{}}MAYO and FNUSA \\ (physiology, pathology)\end{tabular} & F1-Score, F2-Score                                                     \\ \midrule
\multirow{3}{*}{Forecasting}                                 & \begin{tabular}[c]{@{}l@{}}\textbf{Long-term}: MAYO,\\ FNUSA (20 patches)\end{tabular}     & \multirow{3}{*}{MAE, MSE}                                              \\ \cmidrule{2-2}
                                                             & \begin{tabular}[c]{@{}l@{}}\textbf{Short-term}: MAYO,\\ FNUSA (17 patches)\end{tabular}    &                                                                        \\ \midrule
Imputation                                                   & \begin{tabular}[c]{@{}l@{}}MAYO, FNUSA \\ (15 patches)\end{tabular}               & MAE, MSE                                                                        \\ \bottomrule
\end{tabular}
\label{benchmark}
\end{table}

\begin{table*}[ht]
\caption{Model Performance (\%) on Seizure Classification and Seizure Detection}
\centering
\begin{tabular}{cclcccccccccc}
\toprule
\multirow{4}{*}{\begin{tabular}[c]{@{}c@{}}Model\\ Type\end{tabular}}             & \multirow{4}{*}{\begin{tabular}[c]{@{}c@{}}Model\\ Size\end{tabular}} & \multirow{4}{*}{\diagbox{Method}{Task \& \\Dataset}} & \multicolumn{10}{c}{Seizure Classification}                                                                                                                                                                                                       \\ \cmidrule{4-13} 
                                                                                  &                                                                       &                         & \multicolumn{5}{c}{MAYO}                                                                                                       & \multicolumn{5}{c}{FNUSA}                                                                                        \\ \cmidrule(r){4-8} \cmidrule(r){9-13} 
                                                                                  &                                                                       &                         & Acc.                               & Prec.                & Rec.                 & F1                   & F2                   & Acc.                 & Prec.                & Rec.                 & F1                   & F2                   \\ \midrule
\multirow{2}{*}{\begin{tabular}[c]{@{}c@{}}Time\\ Series\end{tabular}}            & 1.36M                                                                 & PatchTST                & 64.17                              & 62.09                & 64.17                & 59.96                & 62.34                & 63.12                & 54.63                & 63.12                & 55.33                & 59.07                \\
                                                                                  & 19.23M                                                                & TimesNet                & 80.80                              & 80.69                & 80.80                & 80.59                & 80.68                & 78.75                & 77.33                & 78.75                & 77.30                & 78.01                \\ \midrule
\multirow{5}{*}{\begin{tabular}[c]{@{}c@{}}EEG\\ Foundation\\ Model\end{tabular}} & 43.18M                                                                & BrainBERT               & 77.11                              & 77.07                & 77.11                & 77.00                & 77.05                & 79.44                & 78.93                & 79.44                & 78.26                & 78.79                \\
                                                                                  & \multirow{2}{*}{79.53M}                                               & Neuro-GPT (Probe)       & 66.11                              & 65.92                & 64.99                & 65.38                & 63.73                & 58.30                & 59.43                & 57.66                & 58.30                & 57.78                \\
                                                                                  &                                                                       & Neuro-GPT (Finetune)    & 85.82                              & 85.94                & 85.82                & 85.83                & 85.82                & 79.83                & 80.08                & 79.83                & 79.87                & 79.82                \\
                                                                                  & \multirow{2}{*}{46M}                                                  & LaBraM (Probe)          & 77.20                              & 77.20                & 77.13                & 77.20                & 76.07                & 77.13                & 77.33                & 76.99                & 76.94                & 76.99                \\
                                                                                  &                                                                       & LaBraM (Finetune)       & \textbf{87.73}                     & \uwave{86.73}               & \textbf{87.80}       & {\ul 87.26}          & {\ul 87.58}          & {\ul 84.94}          & {\ul 84.94}          & {\ul 84.98}          & {\ul 84.96}          & \textbf{84.97}       \\ \midrule
\multirow{4}{*}{Ours}                                                             & \multirow{2}{*}{476.3M}                                               & FoME-Base (Probe)       & 76.65                              & 76.95                & 76.65                & 76.26                & 76.37                & 75.69                & 75.63                & 75.69                & 75.66                & 75.68                \\
                                                                                  &                                                                       & FoME-Base (Finetune)    & \uwave{86.35}                             & {\ul 86.92}          & \uwave{86.35}               & \uwave{86.63}               & \uwave{86.46}               & \uwave{82.22}               & \uwave{82.09}               & \uwave{82.22}               & \uwave{82.12}               & \uwave{82.18}               \\
                                                                                  & \multirow{2}{*}{744.8M}                                               & FoME-Large (Probe)      & 78.65                              & 78.95                & 78.65                & 78.80                & 78.71                & 77.52                & 77.36                & 77.52                & 77.44                & 77.49                \\
                                                                                  &                                                                       & FoME-Large (Finetune)   & {\ul 87.71}                        & \textbf{87.89}       & {\ul 87.68}          & \textbf{87.78}       & \textbf{87.72}       & \textbf{85.52}       & \textbf{85.36}       & \textbf{85.52}       & \textbf{85.44}       & {\ul 84.49}          \\ \midrule
\multirow{4}{*}{\begin{tabular}[c]{@{}c@{}}Model\\ Type\end{tabular}}             & \multirow{4}{*}{\begin{tabular}[c]{@{}c@{}}Model\\ Size\end{tabular}}                                           & \multirow{4}{*}{\diagbox{Method}{Task \& \\Dataset}} & \multicolumn{10}{c}{Seizure Detection}                                                                                                                                                                                                            \\ \cmidrule{4-13} 
                                                                                  &                                                                       &                         & \multicolumn{5}{c}{MAYO}                                                                                                       & \multicolumn{5}{c}{FNUSA}                                                                                        \\ \cmidrule(r){4-8} \cmidrule(r){9-13} 
                                                                                  &                                                                       &                         & Acc.                               & Prec.                & Rec.                 & F1                   & F2                   & Acc.                 & Prec.                & Rec.                 & F1                   & F2                   \\ \midrule
\multirow{2}{*}{\begin{tabular}[c]{@{}c@{}}Time\\ Series\end{tabular}}            & 1.36M                                                                 & PatchTST                & 84.10                              & 84.99                & 84.10                & 80.35                & 82.16                & 87.54                & 87.45                & 87.54                & 87.40                & 87.46                \\
                                                                                  & 19.23M                                                                & TimesNet                & {\ul 94.57}                        & \uwave{94.56}               & {\ul 94.57}          & \uwave{94.57}               & {\ul 94.57}          & 86.10                & 86.22                & 86.10                & 86.13                & 86.10                \\ \midrule
\multirow{4}{*}{\begin{tabular}[c]{@{}c@{}}EEG\\ Foundation\\ Model\end{tabular}} & 505.7M                                                                & Brant                   & 89.40                              & 58.78                & 70.80                & 64.23                & 68.02                & 83.51                & 83.60                & 54.18                & 65.75                & 58.28                \\
                                                                                  & 43.18M                                                                & BrainBERT               & 92.51                              & 92.48                & 92.51                & 92.49                & 92.50                & \uwave{89.85}               & \uwave{89.80}               & \uwave{89.85}               & \uwave{89.77}               & \uwave{89.80}               \\
                                                                                  & 79.53M                                                                & Neuro-GPT               & \uwave{94.39}         & 94.33                & \uwave{94.39}               & 94.35                & \uwave{94.37}               & 88.13                & 88.16                & 88.13                & 88.10                & 88.11                \\
                                                                                  & 46M                                                                   & LaBraM                  & 94.20                              & \textbf{95.21}       & 94.20                & \textbf{95.21}       & 94.20                & {\ul 90.70}          & {\ul 90.71}          & {\ul 90.70}          & {\ul 90.70}          & {\ul 90.70}          \\ \midrule
Ours                                                                              & 744.8M                                                                & FoME                    & {\textbf{95.16}} & {\ul 95.10}          & \textbf{95.16}       & {\ul 95.13}          & \textbf{95.15}       & \textbf{91.80}       & \textbf{91.82}       & \textbf{91.80}       & \textbf{91.81}       & \textbf{91.80}       \\ 
\bottomrule
\end{tabular}
\label{results_seizure}
\end{table*}

\subsubsection{Classification}

Electroencephalogram (EEG) signal classification plays a crucial role in identification and medical diagnostics. We focused on four primary classification tasks:
a) Seizure detection is the most commonly used diagnostic and treatment method, which involves identifying seizure events from continuous patient recordings. We utilized physiological and pathological signals extracted from the MAYO and FNUSA datasets for this binary classification task.
b) Seizure classification: After removing power line interference, we conducted a three-category classification (physiological, pathological, and artifact) using the complete MAYO and FNUSA datasets.
c) Emotion recognition, which aids researchers in understanding brain dynamics, was performed using the SEED dataset for a three-category classification (positive, negative, and neutral). While the SEED dataset provides preprocessed high-quality signal features (e.g., differential entropy, DCAU). Most of the existing work tends to use these provided well-featured data for classification tasks, which limits further exploration of emotional data. While extracting good frequency-domain features can lead to improved classification performance, the utilization of temporal signals is equally important, and, signal data has significant advantages in real-time tasks; therefore, instead of using the preprocessed features provided by the authors of the dataset, we conducted comparative experiments using raw EEG signals to follow the necessary preprocessing procedure adopted in this paper, so as to provide a fair picture of the performance of the various types of models on the true performance on emotional data.
d) Sleep stage identification, crucial for preventing and diagnosing related disorders, was conducted using the SleepEDFx dataset for binary classification.
All classification tasks maintained consistent preprocessing methods with the pre-training phase and employed a 6:2:2 ratio for train-validation-test set partitioning. Evaluation metrics included accuracy, precision, recall, F1 score, and F2 score.

\subsubsection{Forecasting}
EEG signals present unique challenges for forecasting due to their low amplitude (microvolts), complexity, randomness, and transient nature. Despite these difficulties, accurate EEG prediction is crucial for various clinical applications, including disease diagnosis. We conducted comprehensive comparative tests for both short-term and long-term forecasting tasks using the MAYO and FNUSA datasets. For a signal $X=[x_1,...,x_L]$ where $x_i \in \mathbb{R}$, the objective was to predict the subsequent $T$ time steps $[x_{L+1},...,x_{L+T}]$. The past signal length was set to 15 patches (90s) for both tasks, with short-term prediction requiring forecasting of two future patches (12 seconds) and long-term prediction requiring five future patches (30 seconds). Datasets were split into 6:2:2 ratios for training, validation, and testing. Evaluation metrics included Mean Absolute Error (MAE) and Mean Squared Error (MSE).

\subsubsection{Imputation}
EEG signal imputation is essential in real-world applications where data loss can occur due to various factors such as subject movement or equipment malfunction. For a signal $X=[x_1,...,x_L]$ and mask $M=[m_1,...,m_L]$, where $m_i=0$ if $x_i$ is missing and $m_i=1$ if $x_i$ exists, the imputation task involves predicting missing values using existing ones. We considered a patch as valid data only when all values within it existed. For other patches, we employed a learnable mask embedding [MASK] and utilized FoME's default reconstruction head for prediction. The MAYO and FNUSA datasets were split into 6:2:2 ratios for training, validation, and testing to evaluate the model's imputation capabilities. Evaluation metrics included Mean Absolute Error (MAE) and Mean Squared Error (MSE).

\subsubsection{Baselines}
We compared FoME with state-of-the-art deep learning models across various tasks. These included four Transformer-based pre-trained models: BrainBERT (2023) \cite{wang2023brainbert}, Brant (2023) \cite{zhang2024brant}, Neuro-GPT (2024) \cite{cui2023neuro}, and LaBraM (2024) \cite{jiang2024large}; the classic RNN-based model LSTM (1997) \cite{hochreiter1997long}; the CNN-based SOTA model ConvNeXt (2022) \cite{liu2022convnet}; and two advanced temporal models: PatchTST (2023) \cite{nie2022time} and TimesNet (2023) \cite{wu2022timesnet}. For Neuro-GPT and LaBraM, which provided pre-trained parameters, we conducted both probe and fine-tuning experiments to thoroughly validate their performance. In total, we reproduced and tested 8 baseline models for comprehensive comparison.

\subsection{Evaluation Results}
\begin{figure}[h]
\centering
\includegraphics[width=0.9\columnwidth]{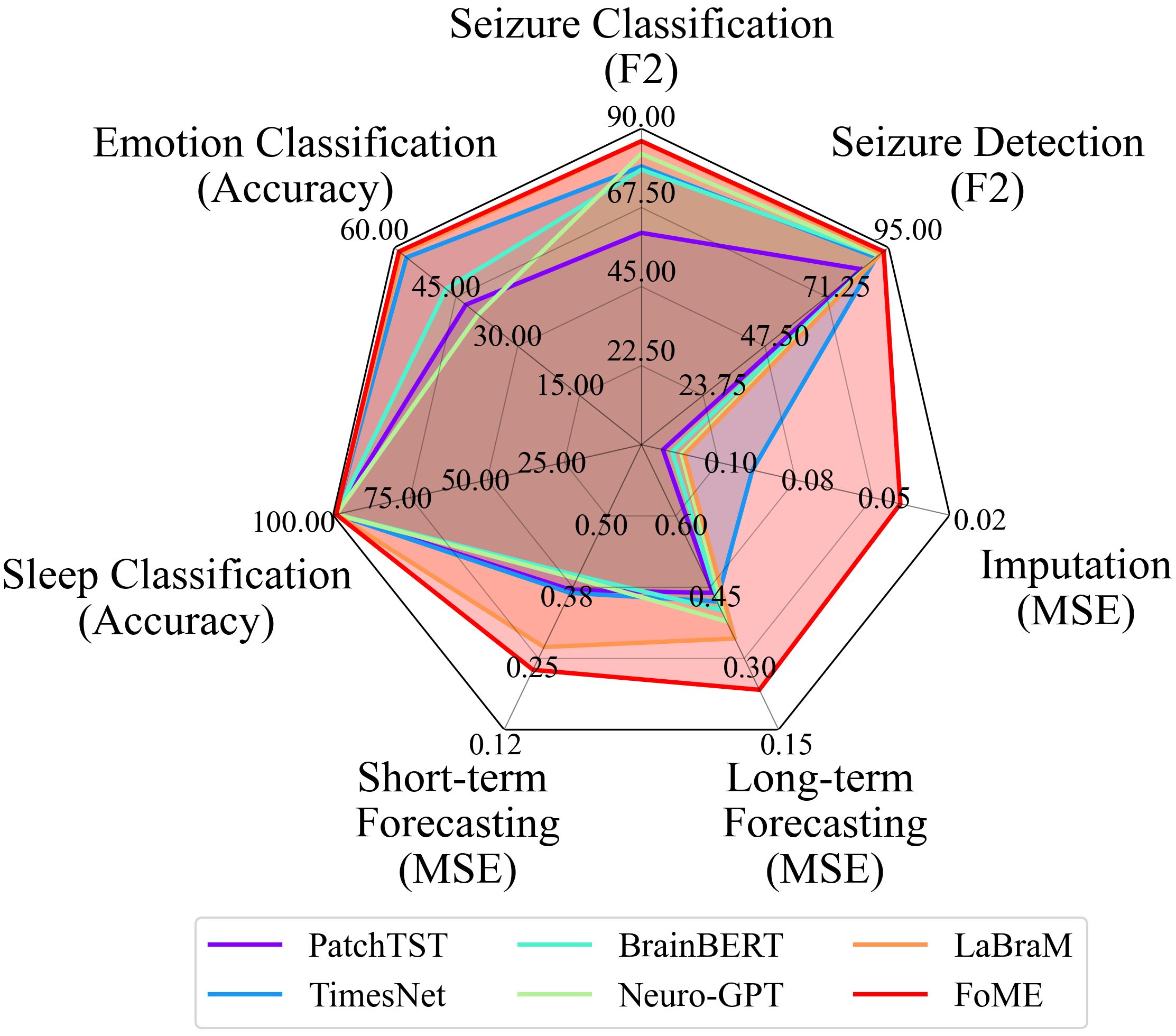}
\caption{The average performance comparison of FoME and other baselines on all downstream datasets.}
\label{performance}
\end{figure}

Fig. \ref{performance} presents the average performance of FoME and the baseline methods across various downstream tasks. It is evident that FoME outperforms existing pre-trained EEG models and time series models on all tasks, with particularly pronounced advantages in signal forecasting and imputation. These results underscore the effectiveness of our pre-training strategy and the model's robust generalization capabilities. In the following sections, we delve deeper into the experimental results, providing detailed analyses for each of the three primary task categories.

\subsubsection{Classification Task }

\paragraph{Seizure Classification \& Seizure Detection}

Table \ref{results_seizure} illustrates the classification performance of FoME and baseline models on epilepsy-related tasks. Given the importance of recall in tasks where false negatives carry high costs, such as seizure detection, we focused our analysis on the F2 score. Across the four subtasks presented in Table \ref{results_seizure}, FoME achieved superior F2 scores in three, surpassing LaBraM by an average of 1.71\%. Several noteworthy observations emerged from this analysis:
1) In binary classification tasks of lesser complexity, inter-model performance disparities were relatively modest. FoME consistently outperformed other models, with TimesNet, BrainBERT, and LaBraM alternating in second place. However, in the more challenging three-category classification task, FoME and LaBraM substantially outperformed the other models. This outcome underscores the efficacy of large-scale EEG data pre-training in facilitating the acquisition of robust features.
2) A positive correlation between model parameter count and performance was observed, aligning with findings from other domains. This trend emphasizes the critical role of model capacity in learning complex representations.
3) Comparative analysis of probing and fine-tuning techniques for FoME, Neuro-GPT, and LaBraM revealed that pre-trained models had already acquired generalizable representations of EEG signals. Task-specific fine-tuning further enhanced model performance, highlighting the significance of task-specific adaptation.

\paragraph{Emotion Classification \& Sleep Stage Classification}

Table \ref{results_seed} presents the three-category classification performance based on the SEED dataset. The results indicate that emotion recognition using minimally processed temporal data remains challenging. In this context, our proposed FoME outperformed other models across multiple metrics, including accuracy and precision. This superiority suggests that the adaptive temporal-lateral attention scaling mechanism effectively captures subtle emotional states in EEG signals.

Table \ref{results_sleep} displays the results of sleep stage classification tests. In contrast to the SEED dataset, the SleepEDFx dataset presented a lower level of difficulty, with most models achieving high classification efficiency. Notably, TimesNet exhibited exceptional performance on this dataset, potentially due to its compatibility with the two-electrode signal structure, which aligns well with the single-channel data typically emphasized in temporal models.

% \usepackage{multirow}
\begin{table}[h]
\caption{Model Performance (\%) on Emotion Classification}
\centering
\begin{tabular}{llccccc}
\toprule
\multirow{4}{*}{Year} & \multirow{4}{*}{\diagbox{Method}{Task \& \\Dataset}} & \multicolumn{5}{c}{Emotion Classification}                                                  \\ \cmidrule{3-7} 
                      &                         & \multicolumn{5}{c}{SEED}                                                                    \\ \cmidrule{3-7} 
                      &                         & Acc.          & Prec.         & Rec.                      & F1             & F2             \\ \midrule
1997                  & LSTM                    & 44.31         & 43.79         & 44.46                     & 42.77          & 43.39          \\
2022                  & ConvNeXt                & 52.50          & 52.40          & 50.96                     & 51.55          & 51.36          \\ \midrule
2023                  & PatchTST                & 42.41         & 46.40          & 42.41                     & 38.74          & 39.90           \\
2023                  & TimesNet                & 57.13         & 59.72         & 57.13                     & 57.67          & 57.18          \\ \midrule
2023                  & BrainBERT               & 47.56         & 50.19         & 47.56                     & 46.79          & 46.72          \\
2024                  & Neuro-GPT               & 39.73         & 40.20          & 39.73                     & 39.49          & 39.54          \\
2024                  & LaBraM                  & 57.93         & 58.99         & 57.93                     & \textbf{58.99} & 57.99          \\ \midrule
Ours                  & FoME                    & \textbf{58.80} & \textbf{60.30} & \textbf{58.80}             & 57.89          & \textbf{58.06} \\ \bottomrule
\end{tabular}
\label{results_seed}
\end{table}

% \usepackage{multirow}
\begin{table}[h]
\caption{Model Performance (\%) on Sleep Stage Classification}
\centering
\begin{tabular}{llccccc}
\toprule
\multirow{4}{*}{Year} & \multirow{4}{*}{\diagbox{Method}{Task \& \\Dataset}} & \multicolumn{5}{c}{Sleep Stage Classification}                                     \\ \cmidrule{3-7} 
                      &                         & \multicolumn{5}{c}{SleepEDFx}                                                      \\ \cmidrule{3-7} 
                      &                         & Acc.           & Prec.          & Rec.           & F1             & F2             \\ \midrule
1997                  & LSTM                    & 80.03          & 79.34          & 82.33          & 80.74          & 80.00          \\
2022                  & ConvNeXt                & 92.65          & 91.27          & 92.68          & 92.00          & 92.97          \\ \midrule
2023                  & PatchTST                & 98.79          & 97.61          & 98.79          & 98.20          & 98.55          \\
2023                  & TimesNet                & \textbf{99.25} & 98.51          & \textbf{99.25} & 98.88          & \textbf{99.10} \\ \midrule
2023                  & BrainBERT               & 98.80          & 97.61          & 98.80          & 98.20          & 98.56          \\
2024                  & Neuro-GPT               & 98.90          & 97.81          & 98.90          & 98.35          & 98.68          \\
2024                  & LaBraM                  & 98.77          & 97.75          & 98.77          & 98.16          & 98.52          \\ \midrule
Ours                  & FoME                    & 99.03          & \textbf{98.68} & 98.73          & \textbf{99.10} & 98.47         \\ \bottomrule
\end{tabular}
\label{results_sleep}
\end{table}

\subsubsection{Signal Forecasting Task}
Table \ref{results_forecasting} presents the results of short-term and long-term EEG signal forecasting tasks. FoME consistently outperformed other models across all tasks, further validating its efficient modeling of signal sequences and demonstrating precise prediction of future signal trends. This capability has multifaceted significance:
In the medical domain, accurate prediction of EEG trends can facilitate timely warnings in the critical pre-ictal phase of epileptic seizures. In neuroscience research, precise EEG signal prediction can deepen scientists' understanding of brain mechanisms and neuronal activity patterns, enabling further exploration of information processing modalities and neuronal interactions across various brain states.

% \usepackage{multirow}
\begin{table*}[h]
\caption{Model Performance on Signal Forecasting and Imputation}
\centering
\begin{tabular}{llcccccccccccc}
\toprule
\multirow{4}{*}{Year} & \multirow{4}{*}{Method} & \multicolumn{4}{c}{Short-term Signal Forecasting}                     & \multicolumn{4}{c}{Long-term Signal Forecasting}                      & \multicolumn{4}{c}{Imputation}                                        \\ \cmidrule{3-6} \cmidrule{7-10} \cmidrule{11-14}
                      &                         & \multicolumn{2}{c}{MAYO}          & \multicolumn{2}{c}{FNUSA}         & \multicolumn{2}{c}{MAYO}          & \multicolumn{2}{c}{FNUSA}         & \multicolumn{2}{c}{MAYO}          & \multicolumn{2}{c}{FNUSA}         \\ \cmidrule{3-4} \cmidrule{5-6} \cmidrule{7-8} \cmidrule{9-10} \cmidrule{11-12} \cmidrule{13-14}
                      &                         & MAE             & MSE             & MAE             & MSE             & MAE             & MSE             & MAE             & MSE             & MAE             & MSE             & MAE             & MSE             \\ \midrule
2023                  & TimesNet                & 1.0086          & 0.8214          & 0.8101          & 1.1207          & 0.8582          & 1.1616          & 0.8310          & 1.1543          & 0.9207          & 1.3456          & 0.9225          & 1.3942          \\
2023                  & PatchTST                & 0.7360          & 0.9661          & 0.7366          & 0.9640          & 0.8119          & 1.1140          & 0.7998          & 1.0815          & 0.3381          & 0.2655          & 0.3358          & 0.2833          \\ \midrule
2023                  & BrainBERT               & 0.8357          & 1.1210          & 0.8021          & 1.0584          & 0.8100          & 1.0490          & 0.7851          & 1.0089          & 0.7780          & 0.9788          & 0.7791          & 1.0000          \\
2023                  & Brant                   & 0.6860          & 0.8064          & 0.6915          & 0.8249          & -               & -               & -               & -               & 0.4887          & 0.4622          & 0.5020          & 0.4835          \\
2024                  & Neuro-GPT               & 0.7688          & 0.9762          & 0.8289          & 1.1366          & 0.7387          & 0.8969          & 0.7926          & 1.0346          & 0.6316          & 0.7113          & 0.6794          & 0.8056          \\
2024                  & LaBraM                  & 0.5836          & 0.5937          & 0.6896          & 0.8130          & 0.6806          & 0.7866          & 0.7580          & 0.9818          & 0.6034          & 0.6368          & 0.6550          & 0.7500          \\ \midrule
Ours                  & FoME                    & \textbf{0.5675} & \textbf{0.5874} & \textbf{0.6157} & \textbf{0.6796} & \textbf{0.6162} & \textbf{0.6649} & \textbf{0.6446} & \textbf{0.7301} & \textbf{0.1913} & \textbf{0.1196} & \textbf{0.1969} & \textbf{0.1198} \\ \bottomrule
\end{tabular}
\label{results_forecasting}
\end{table*}

\subsubsection{Imputation Task}

\begin{figure*}[t]
\centering
\includegraphics[width=1.0\textwidth]{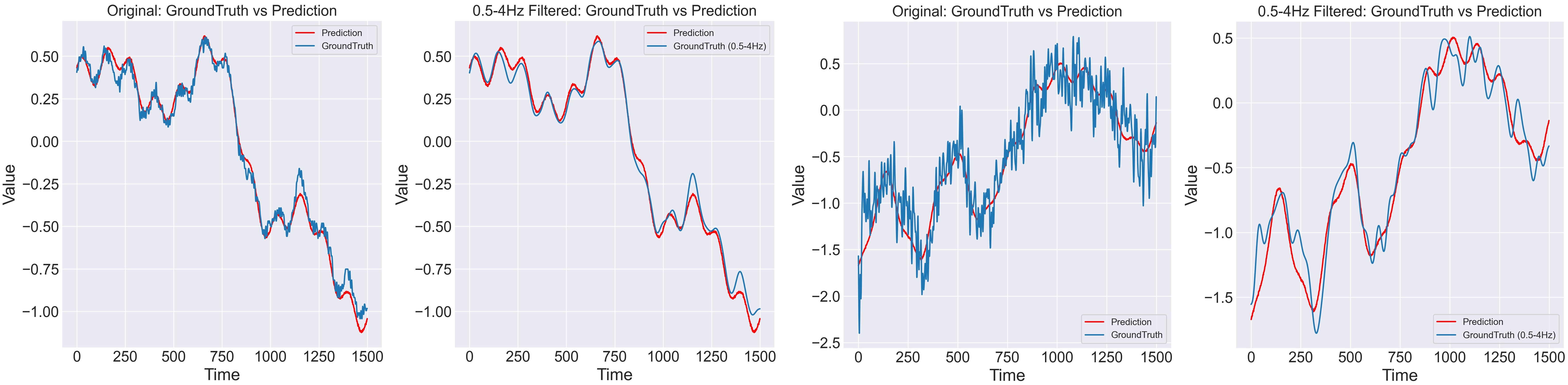}
\caption{Visualization of imputation results given by FoME under the 40\% mask ratio setting.}
\label{imputation}
\end{figure*}

The results of EEG signal imputation tests are presented in Table \ref{results_forecasting}. FoME exhibited exceptional performance in this task, significantly outperforming all other models. Compared to the current state-of-the-art model, PatchTST, FoME achieved an average loss reduction of 0.1488, underscoring its capacity for accurate reconstruction of missing data points.
To provide a more intuitive representation of signal reconstruction efficacy, we selected samples from the test dataset for visualization. As illustrated in Fig. \ref{imputation}, the first column represents the original data, while the second column displays the original data after low-frequency filtering (0.5-4 Hz). The results demonstrate FoME's high accuracy in signal prediction.

\begin{figure}[h]
\centering
\includegraphics[width=1.0\columnwidth]{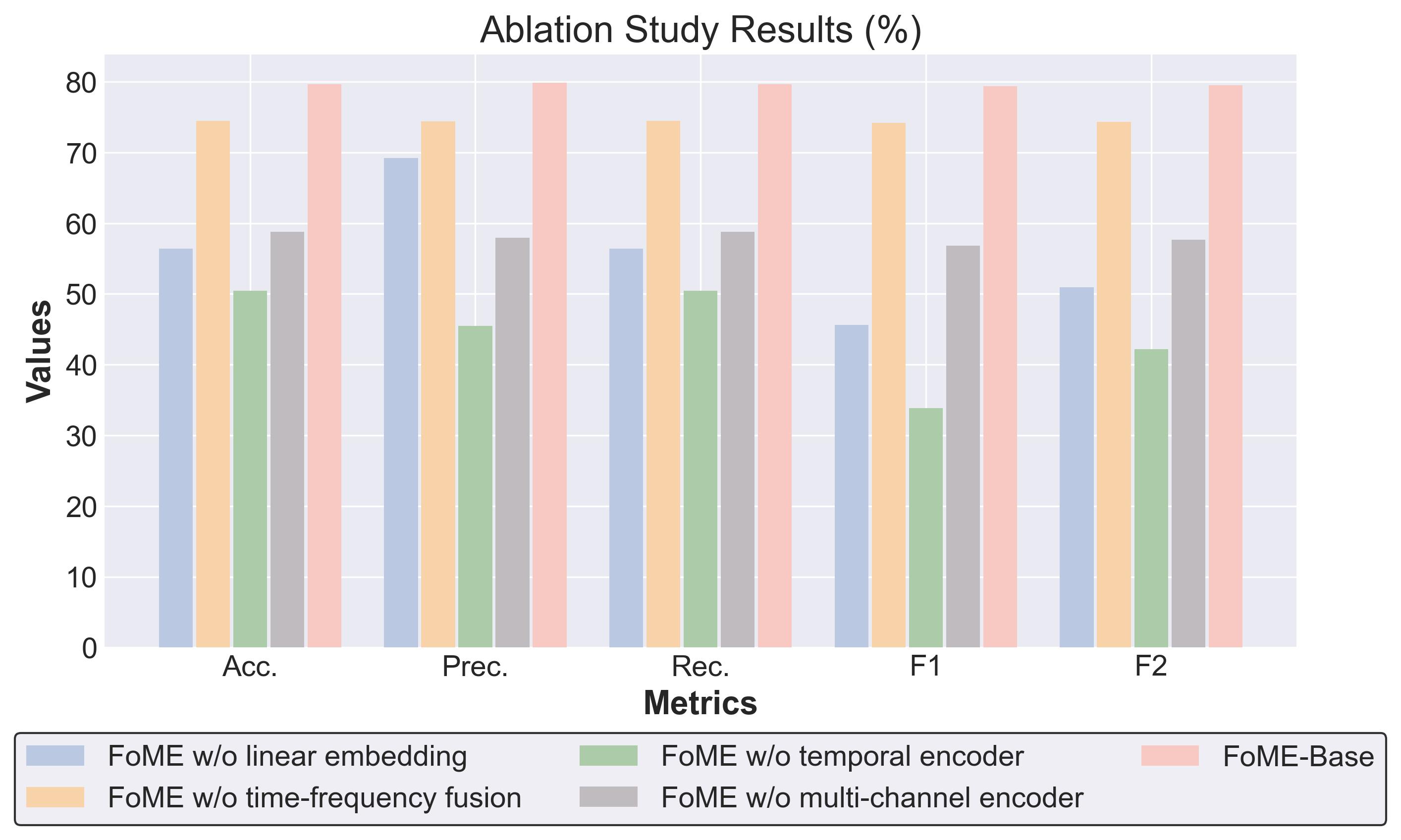}
\caption{Ablation study based on the TUEV dataset.}
\label{ablation}
\end{figure}

\subsection{Ablation Study}

We strongly believe that FoME's superior performance stems from our proposed time-frequency fusion encoding and adaptive temporal-lateral attention scaling. To validate the efficacy of these modules, we implemented the following FoME variants: 1) FoME without time-frequency fusion: This variant omits the encoding and integration of frequency information, relying solely on temporal embedding and positional encoding for token generation.
2) FoME without temporal encoder: Removal of the temporal encoder diminishes the model's compatibility with data featuring fewer electrodes.
3) FoME without multi-channel encoder: Removal of the adaptive multi-channel encoder results in a loss of spatial modeling capabilities.
4) FoME with CNN module replacing linear embedding: This variant substitutes the linear embedding with a CNN module.
We conducted experiments on the TUEV dataset to evaluate these four model variants. Their performance is illustrated in Fig. \ref{ablation}, demonstrating that all four components contribute to FoME-Base's performance enhancement. Notably, the removal of the temporal encoder had the most substantial impact on model performance, affirming its crucial role in signal modeling.

\begin{figure}[h]
\centering
\includegraphics[width=1.0\columnwidth]{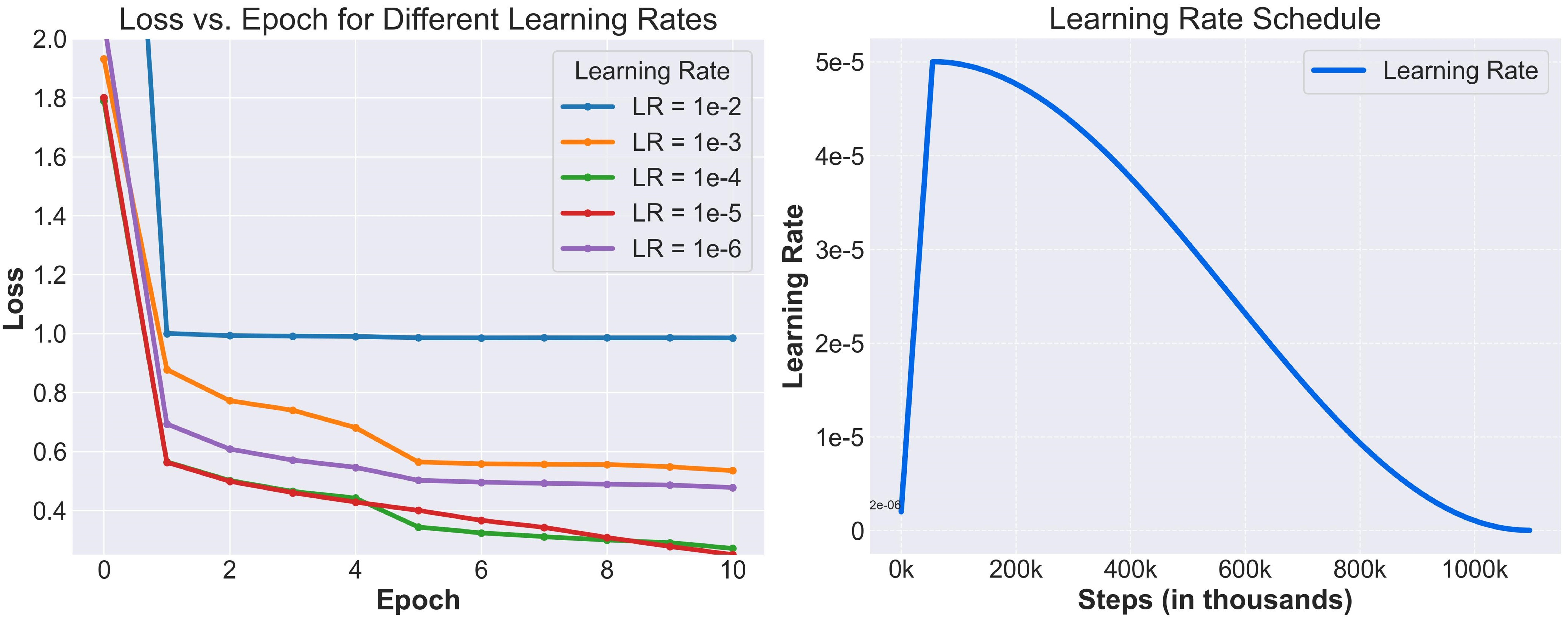}
\caption{Learning rate experiment and scheduling rules.}
\label{learning_rate}
\end{figure}

The configuration and modulation of learning rates are pivotal to model pre-training. Consequently, prior to large-scale training, we conducted a learning rate range test using FoME-Base to identify the optimal learning rate range for our data and model architecture. It is worth noting that larger datasets typically accommodate higher learning rates due to the availability of more samples for gradient stabilization. To minimize discrepancies between the test environment and the final training environment, we selected a substantial dataset of approximately 500 GB for this validation, completing 10 epochs of training.
The experimental results, as depicted in Fig. \ref{learning_rate}-left, indicate that learning rates of $1.0\times 10^{-4}$ and $1.0\times {10^{-5}}$ yielded the lowest loss values and fastest convergence rates. A clear downward trend persisted even after 10 epochs. Based on these findings, we selected $5.0\times 10^{-5}$ as the peak learning rate for large-scale training. The scheduling process is illustrated in Fig. \ref{learning_rate}-right.

\section{Conclusion and Discussion}

\subsection{Conclusion}

We propose FoME, a novel EEG foundation model that addresses the longstanding challenges in EEG signal analysis through the implementation of adaptive temporal-lateral attention scaling. FoME leverages self-supervised learning on a massive and diverse dataset exceeding 1.7TB, encompassing both scalp and intracranial recordings. This pre-training empowers FoME to capture intricate dynamics inherent in EEG signals, facilitating robust and generalizable performance across various downstream tasks. The key innovations of FoME are the time-frequency fusion embedding and the adaptive time-lateral attention scaling (ATLAS) mechanism, which have proven instrumental in capturing the complex temporal and spectral dynamics inherent in EEG signals. These components synergistically enable FoME to adapt to varying patterns across diverse data streams and facilitate robust multi-channel modeling, addressing the heterogeneity challenge that has long plagued EEG analysis. Our comprehensive evaluations across four downstream tasks, including classification, short-term forecasting, long-term forecasting, and signal imputation, consistently demonstrate FoME's superior performance compared to state-of-the-art models. Notably, FoME's exceptional capabilities in signal forecasting underscore its potential for critical applications such as early seizure detection and advanced brain-computer interfaces. FoME represents a significant advancement in EEG analysis, offering a versatile foundation that transcends the limitations of task-specific models. By enabling effective transfer learning and potentially zero-shot capabilities, FoME paves the way for more accessible and powerful EEG-based applications across neuroscience, clinical diagnostics, and brain-computer interfaces.

\subsection{Limitation}
Despite FoME's significant advancements in EEG analysis, several limitations warrant consideration. The pre-training dataset, while substantial, underrepresents certain EEG modalities such as emotion-related, sleep, and motor imagery data, and lacks representation from cognitive domains like working memory tasks. Additionally, the underutilization of invasive EEG data, which typically offers higher signal-to-noise ratios, presents an opportunity for improvement. The current scale of FoME (745M parameters) is constrained by the available pre-training data, suggesting potential for further scaling as the dataset expands. Moreover, while FoME excels in short-term signal prediction, its capabilities for long-term forecasting remain unexplored.

Future research directions should focus on addressing these limitations. Expanding and diversifying the pre-training dataset to include a wider range of EEG modalities and cognitive tasks is crucial. Furthermore, refining preprocessing strategies specific to EEG signals and exploring more sophisticated pre-training techniques, including the integration of decoder architectures, could potentially improve the model's generalization capabilities and downstream task performance. Finally, future iterations should explore methods to increase the input sequence length beyond the current 90-second limit, potentially improving performance on tasks requiring extended temporal context.

\section*{Acknowledgments}
This work was supported in part by XXX.


\bibliographystyle{IEEEtran}
\bibliography{main}{}

% 垂直填充空间
\vfill

\end{document}